\documentclass[conference,10pt,a4paper]{IEEEtran}
\usepackage{algorithmic,algorithm,url,amsmath,cite,caption,array}
\ifx\pdfoutput\undefined
  \usepackage[pdftex]{graphicx}
\else
  \usepackage{graphicx}
\fi

\begin{document}

\title{Exhaustive Search-based Model for Hybrid Sensor Network}

\author{\authorblockN{A.A. Waskita\authorrefmark{1}\authorrefmark{2}, 
H. Suhartanto\authorrefmark{2}, Z. Akbar\authorrefmark{3}\authorrefmark{4},
L.T. Handoko\authorrefmark{3}\authorrefmark{5}}
\authorblockA{\authorrefmark{1}Center for Development of Nuclear
Informatics, National Nuclear Energy Agency, \\
Kawasan Puspiptek Serpong, Tangerang 15310, Indonesia\\
Email : adhyaksa@batan.go.id}
\authorblockA{\authorrefmark{2}Faculty of Computer Science, University of
Indonesia, \\
Kampus UI Depok, Depok 16424, Indonesia\\
Email : heru@cs.ui.ac.id}
\authorblockA{\authorrefmark{3}Group for Theoretical and Computational Physics,
Research Center for Physics, Indonesian Institute of Sciences, \\
Kawasan Puspiptek Serpong, Tangerang 15310, Indonesia\\
Email: zaenal@teori.fisika.lipi.go.id, handoko@teori.fisika.lipi.go.id}
\authorblockA{\authorrefmark{4}Group for Bioinformatics and Information
Mining, Department of Computer and Information Science, \\
University of Konstanz, Box D188, D-78457 Konstanz, Germany\\
Email: zaenal.akbar@uni-konstanz.de}
\authorblockA{\authorrefmark{5}Department of Physics, University of Indonesia,\\
Kampus UI Depok, Depok 16424, Indonesia\\
Email: handoko@fisika.ui.ac.id}
}

\maketitle

\begin{abstract}
A new model for a cluster of hybrid sensors network with multi
sub-clusters is proposed. The model is in particular relevant to the
early warning system in a large scale monitoring system in, for example, a
nuclear power plant. It mainly addresses to a safety critical system
which requires real-time processes with high accuracy. The mathematical model is
based on the extended conventional search algorithm with certain interactions
among the nearest neighborhood of sensors. It is argued that the model could
realize a highly accurate decision support system with less number of
parameters. A case of one dimensional interaction function is discussed, and
a simple algorithm for the model is also given.
\end{abstract}

\section{Introduction}
\label{sec:intro}

A safety critical system in, for example, nuclear power plants involves
complicated and advance safety system for ensuring its running process
absolutely safe. The system consists of high number and various types
of sensors \cite{jamilAffandi,SciTopics}. All of them generate
a huge amount of data at a real-time basis which must be processed properly
throughout its whole life time. Many heuristic approaches based on the
artificial intelligence (AI) such as neural network are currently available and
have been studied intensively \cite{Obreja,waveletRBNN,recurrentNN,pakDede}.
However, the AI-based approaches have a fundamental problem due to its
statistical algorithm which could lead to disaster in the real applications of
safety critical system. 

It is clear that in such critical systems, no fault tolerance is the most
important principle. Therefore, putting the safety as the priority, one should
implement the exhaustive algorithm spanning over all possibilities
rather than using the AIs. This turns into the exhaustive search problem which
unfortunately lacks of inefficiency due to the requirement of huge computing
power. Some technical approaches have been introduced to overcome this problem.
Most of them deploy the parallel algorithm \cite{Karnin1984,Hui2010} together
with graphical or combinatorial representation to improve both resources and
running time \cite{Matwin1985,haystack,Bhalch2010,Chang2011}.

Previously, the application of exhaustive methods into search problem was not
feasible for large number of sensors. Fortunately, the affordable parallel
environment using graphic processor unit (GPU) is available in recent days.
 The use of GPU is getting popular, especially, after the introduction of the
NVIDIA Compute Uniﬁed Device Architecture (CUDA) through a C-based API
\cite{cuda}. This enables an easy way to take advantage of the high performance
of GPUs for parallel computing. Deploying the GPU-based distributed computing
would reduce the execution time causing less responsive system in the previous
days, while it also realizes lower power and space consumption than CPU
\cite{knapsackGPU}. This motivates us to reconsider the feasibility
of exhaustive search for hybrid sensors network. 

In this paper a new exhaustive search based model is proposed. The model is 
mainly intended to realize an exhaustive decision support system (DSS)
consisting of various and huge number of sensors. However, the paper is focused
only on introducing the model. The discussion of parallelization and detail
analysis will be published elsewhere.

The paper is organized as follows. First in Sec. \ref{sec:model} the model is
introduced, and it is followed by the description of mathematical formulation in
Sec. \ref{sec:math}. In Sec. \ref{sec:alg} a simple algorithm to execute the
model is given. Finally, the paper is ended with a short summary and discussion.

\section{The model}
\label{sec:model}

Before moving on constructing the model and its mathematical representation,
let us discuss the basic constraints and circumstances in the expected
applications. Putting in mind that the model is developed under the following
considerations :
\begin{enumerate}
\item \textbf{No failure decision is allowed.}\\
By definition, there is no room for even a small mistake generated by the DSS.
This actually discourages the deployment of any AI-based method from first
principle.
\item \textbf{Fast enough 'real-time' process}\\
Fast execution time of the whole process is crucial to increase the safety.
However, fast execution time at the order of minutes is in practical more than
enough. This moderate requirement encourages the implementation of exhaustive
methods supported by GPU powered computation.
\item \textbf{Huge number of hybrid sensors}\\
The system is consisting of huge number (at the order of
hundreds or thousands) sensors with different characteristics, in
particular types and scales \cite{domainSafetyAnalysis}.
\item \textbf{Certain relationship across the sensors}\\
Each sensor has  relations with another ones in certain way at some 
degrees. The relationships among the sensors are thereafter called as
interaction. 
\item \textbf{Sensor network with multi clusters}\\
The sensor network is divided into several clusters which typically represents
the geographical locations with different degrees of interaction. This
realizes a situation of, for instance, a nuclear power plant which is equipped
with many sensors in each building. Consequently the sensors in each cluster
have stronger interactions among themselves than with another ones belong to
another clusters. So, the model should be able to describe the independent
interactions among the sensors in a cluster, and also the interactions among
different clusters as well.
\item \textbf{Dynamic behavior}\\
The values of each sensor are by nature changing from time to time. However,
the data acquisition is performed periodically, for instance every few minutes
according to the above second point. In a nuclear reactor facility this could
happen due to human errors, common system failures and even seismic activities
\cite{binarydecission}. 
\end{enumerate}

\begin{figure}[t!]
 \centering
 \includegraphics[width=85mm]{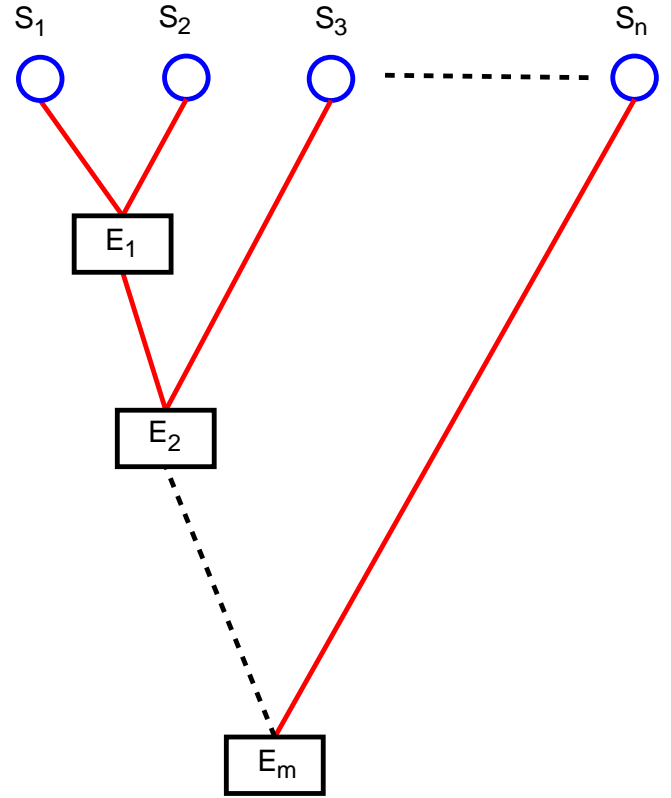}
\caption{The illustration of 1-D relationship of $n$ sensors evaluated till the
$m^\mathrm{th}$ level, where $S$ denotes sensor and $E$ is the evaluation
result.}
\label{fig:simpleEval}
\end{figure}

Having the above requirements in mind, obviously one will arrive at the problem
of unlimited decision trees. In order to reduce the tree significantly without
raising the risk of failures, let us assume the nearest neighborhood
approximation (NNA). Under this approximation, only the interactions with the
nearest sensors are taken into account. 

\begin{figure}[t!]
 \centering
 \includegraphics[width=85mm] {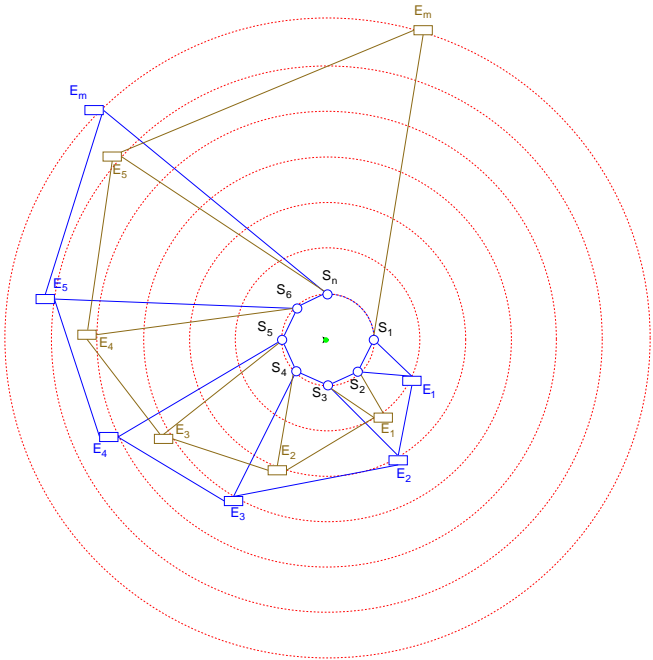}
\caption{The circular representation for 1-D relationship of $n$ sensors evaluated for two schemes (blue and brown lines) till the $m^\mathrm{th}$ level, where $S$ denotes sensor and $E$ is the evaluation result.}
\label{fig:representasi}
\end{figure}

In the present paper, let us consider the simplest case of one dimensional
(1-D) relationship. This means all sensors are put on a virtual line which
allows only interactions with the nearest right and left neighboring
sensors for each sensor. This actually reproduces the known tree analysis
commonly implemented in the analysis of fault
\cite{ftaAircraft,binarydecission}, elements interaction 
\cite{Park_parallel,noisyGraph, socialNet} and even in optimizing system
\cite{airportLayout}. 

The algorithm can be well illustrated in a tree-like diagram using the
evaluation scheme under the NNA scheme as shown in Fig. \ref{fig:simpleEval}. In
the figure, two adjacent sensors are first evaluated and then the result are
subsequently evaluated with another adjacent sensor. The evaluation scheme 
depicted in Fig. \ref{fig:simpleEval} can be exhaustively changed
according to the acquired values from the responsible sensors. It should be
noted that the evaluation scheme is not necessarily binary, but it could be
anything else like fuzzy and so on.

On the other, due to point 4 one should consider the modified tree analysis,
that is both edges should interact each other too and forms a circle line of
sensors. Moreover, each sensor on the circle line should be put carefully.
Because, according to point 5 and the NNA, the relative location of sensors on
the circle line represents their degree of relationship or relevancy between
one and another. The stronger relationship between two sensors, both should be
put closer each other. This type of circular model is depicted in Fig.
\ref{fig:representasi}. There are two examples of evaluation results in the
figure, the blue and brown ones corresponding to the evaluation at different
time. The innermost circles represent the chain of sensors, and the subsequent
outer circles describe the evaluation results at certain levels.

\begin{figure}[b!]
 \centering
 \includegraphics[width=85mm] {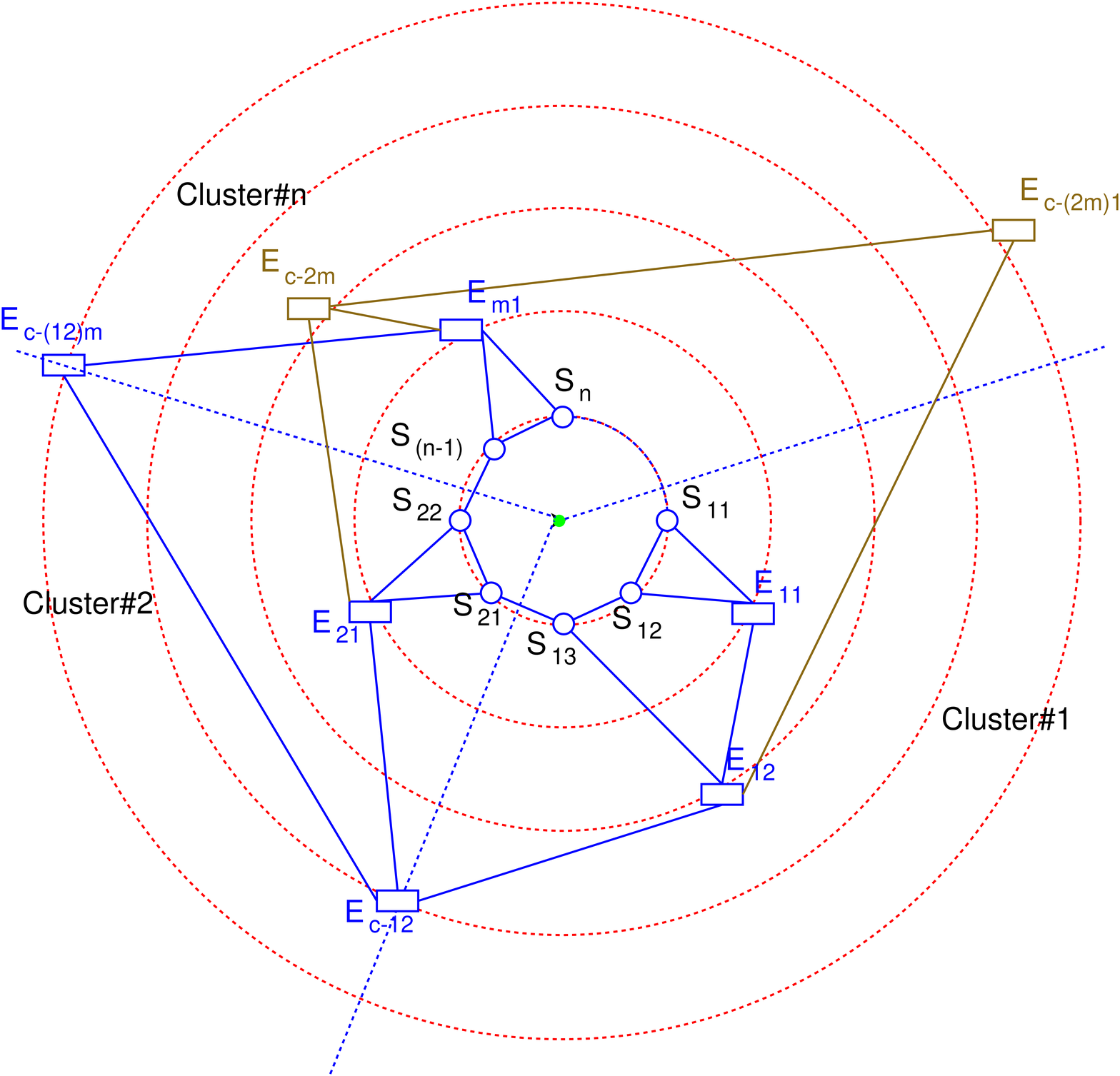}
\caption{The circular representation for 1-D relationship of $n$ sensors belong
to separated clusters, evaluated for two schemes (blue and brown lines) till the
$m^\mathrm{th}$ level, where $S$ denotes sensor and $E$ is the evaluation
result.}
\label{fig:representasiCluster}
\end{figure}

As required in point 5, the sensor network should also be divided into
several clusters based on either its genuine characteristic or critical 
levels. Each cluster can be treated separately as an independent sensor network
as Fig. \ref{fig:representasi}. The model with several clusters is illustrated
in Fig. \ref{fig:representasiCluster} where each cluster is separated by the
blue dashed lines. Two evaluation results are shown in the figure as before, the
blue and brown ones corresponding to the evaluation at different time.

Now, one is ready to formulate the model in a mathematical representation. 

\section{Mathematical representation}
\label{sec:math}

Based on the previous discussion, one should first model the interaction
between the neighboring sensors. This should be a function which determines the
value of $E_{i,j}$ representing the interacting result between two
adjacent sensors at the point $i$ and $j$,
\begin{equation}
 E_{ij} = a_{ij} \, \left( \frac{x_i}{l_i + 1} + \frac{x_j}{l_j + 1}
\right) \; .
\label{eq:interaction}
\end{equation}
Here, $l_i = 1, 2, \cdots, m$ denotes the evaluation level of $x_i$ with $i
= 1,2, \cdots, n$ and $m = n - 1$. While $x_i$ is the normalized value acquired
by sensor for $l_i = 1$, or the value of previous evaluation result for $l_i >
1$. The coupling constant $a_{ij}$ reflects the strength of interaction between
both adjacent sensors, $x_i$ and $x_j$ respectively. It is defined in a way such
that the value of $E_{ij}$ is getting smaller for higher evaluation level, that
is,
\begin{equation}
  a_{ij} = 1 - \frac{| i - j |}{n} \; ,
  \label{eq:aij}
\end{equation}
for $l_i = 1$ in a system with $n$ sensors.

Moreover, Eqs. (\ref{eq:interaction}) and (\ref{eq:aij}) can be extended for
$l_i > 1$ as follow,
\begin{equation}
  E_{(ij)k} = a_{ijk} \, \left( \frac{x_{ij}}{l_{ij} + 1} + \frac{x_k}{l_k +
1} \right) \; ,
  \label{eq:interaction2}
\end{equation}
with,
\begin{equation}
  a_{ijk} = \frac{a_{ij} + a_{ik} + a_{jk}}{3} \; ,
  \label{eq:aij2}
\end{equation}
and $x_{ij} = E_{ij}$ respectively. The situation is well illustrated in
Figs. \ref{fig:representasi} and \ref{fig:representasiCluster}. Further
generalization of Eqs. (\ref{eq:interaction2}) and (\ref{eq:aij2}) up to
certain evaluation level can be performed in a straightforward way.

\begin{table}[t!]
\begin{center}
\caption{The values of weight parameter $a_{ij}$ for $n$ sensors.}
\label{tab:aij}
\begin{tabular}{c|cccccccc}
$a_{ij}$ & 1 & 2 & 3 &  & $\cdots$ & & $n - 1$ & $n$ \\
\\
\hline
\\
1 & 1 &  $\frac{n-1}{n}$ &  $\frac{n-2}{n}$ &  & $\cdots$ & & $\frac{2}{n}$ & 
$\frac{1}{n}$\\
2 & $\frac{n-1}{n}$ & 1  & $\frac{n-1}{n}$  &  & $\cdots$ & & $\frac{3}{n}$ &
$\frac{2}{n}$\\
3 & $\frac{n-2}{n}$ &  $\frac{n-1}{n}$ & 1  &  & $\cdots$ & & $\frac{4}{n}$ &
$\frac{3}{n}$ \\
$\vdots$ & $\vdots$ &  $\vdots$ &  $\vdots$  &  & 1 &  & $\vdots$ & $\vdots$\\
$n-1$ & $\frac{2}{n}$ & $\frac{3}{n}$  & $\frac{4}{n}$ &  & $\cdots$ &  & 1 &
$\frac{n-1}{n}$\\
$n$ & $\frac{1}{n}$ & $\frac{2}{n}$ & $\frac{3}{n}$ &  & $\cdots$ & &
$\frac{n-1}{n}$  & 1 \\
\end{tabular}
\end{center}
\end{table}

There is actually an important reason for choosing the definition in Eqs.
(\ref{eq:aij}) and (\ref{eq:aij2}). In order to fulfill the requirements in
point 3 and 4 in Sec. \ref{sec:model}, it is plausible to normalize all scales
in a uniform unit scale. In the present case all values are normalized to be in
the range between 0 and 1. This normalization demands all acquired values from
the sensors should also be normalized accordingly through a relation,
\begin{equation}
  x_i = f \, \left( x^\prime_i - x_\mathrm{min} \right) \; ,
\end{equation}
where $x_i$ and $x^\prime_i$ are the normalized and originally acquired values
of $i^\mathrm{th}$ sensor for $l_i = 1$. The normalization factor $f$ is,
\begin{equation}
  f = \frac{1}{\left| x_\mathrm{max} - x_\mathrm{min} \right|} \; ,
\end{equation}
with $x_\mathrm{max/min}$ denotes the maximum or minimum value of each sensor.
This kind of normalization enable us to treat all sensors in the same manner
regardless with its types and unit scales.

From Eqs. (\ref{eq:aij}) and (\ref{eq:aij2}), it is obvious that the coupling
$a$ is uniquely characterizing the present model. It ensures the evaluation
value of $E$ at the final $(n - 1)^\mathrm{th}$ level is always divergent, that
is between 0 and 1. This is in contradiction with any conventional tree analysis
which associates the largest evaluation value at the final level as the final
solution. By the way, from the definition $1/n \le a_{ij} \le 1$ as shown in
Tab. \ref{tab:aij} which forms a symmetric matrix with unit diagonal elements.

Furthermore, one should take a threshold value $E_\mathrm{th}$ as a standard
value whether the evaluation value at certain level is allowed to proceed
further or not. For the sake of simplicity, this value is fixed and valid for
all levels and sensors. This represents the critical value of safety. Following
the above normalization it is again constrained,
\begin{equation}
  0 < E_\mathrm{th} < 1 \; .
  \label{eq:eth}
\end{equation}
Only if the evaluation value exceeds this threshold, i.e. $E > E_\mathrm{th}$,
the tree should be analyzed further. Otherwise it ends forever. According to the
initial value of each sensor at certain time, some evaluation values at final
level may survive or not. The surviving value triggers the warning alarm which
is indicating some anomalies detected by any sensors. Of course,
the determination of appropriate $E_\mathrm{th}$ requires preliminary
experiments based on the available standards and regulations. 

The above procedure is carried out each time following the periodic data
acquisition by all sensors. Finally, all the tools have been established and we
are ready for applying the above rules.

\section{The algorithm}
\label{sec:alg}

In this section, let us provide a simple algorithm to realize the previously
discussed model. 

Because of the exhaustive method deployed in the model, it requires all
variables are treated as a circle. The array of variables should be a traversing
pointer indicating the first element as a starting point. Further, the
evaluation is performed from the starting point, proceeds to the
subsequent element of variables array in an increasing mode till reaching the
last one, that is back to the first element. Each sensor has a chance to become 
a root of a tree and also a leaf.

While forming a circle model, the tree within the model is evaluated recursively. 
A set of simple algorithms is presented here. It consists of two main parts for the pra-evaluation and 
the main evaluation. Each sensor is labeled with an integer ranging from $0$ to $n-1$ for $n$ 
number of sensors. The algorithm starts with positioning the sensors by shifting it one by one to 
generate considerable combinations. Each time a tree of sensors is formed, it is evaluated 
as a scheme depicted in Fig. \ref{fig:simpleEval}. 

Algorithm \ref{alg:ps}-\ref{alg:value} require the array of sensor's indexes and values.
The value described in a sensor's index array absolutely points to its value
in the array of sensor's value. Algorithm \ref{alg:ps} determines the positioning of sensors, 
while Algorithm \ref{alg:weight} and \ref{alg:value} evaluate their interactions.

\begin{algorithm}
\caption{Positioning sensors}
\label{alg:ps}
\begin{algorithmic}[1] 
\REQUIRE $root$
\COMMENT{index of sensor being a root of a tree or a sub tree}
\REQUIRE $n$
\COMMENT{number of sensors involved}
\REQUIRE $index$
\COMMENT{array of sensor's index}
\IF{$n=2$ \OR $root=n-2$}
  \STATE evaluate their interaction
  \STATE exchange sequence of the two or the last two sensors
  \STATE evaluate their interaction
\ELSE
  \FOR {$i=0 \to n$}
    \STATE re-positioning sensors($index$,$root+1$)
    \STATE shifting the sequence from the $root$ position
  \ENDFOR
\ENDIF
\end{algorithmic}
\end{algorithm}

\begin{algorithm}
\caption{Calculating interaction's weight}
\label{alg:weight}
\begin{algorithmic}[1] 
\REQUIRE $root$
\COMMENT{index of sensor being a root of a tree or a sub tree}
\REQUIRE $n$
\COMMENT{number of sensors involved}
\REQUIRE $index$
\COMMENT{array of sensor's index}
\REQUIRE $w=0$
\COMMENT{weight of sensor's interaction initialized}
\REQUIRE $b=0$
\COMMENT{counting number of sensors combination currently involved in interaction}
\FOR{$i=1 \to root$}
  \FOR{$j=i+1 \to root$}
    \STATE $w=w+(1-\frac{abs(index[i]-index[j])}{n})$
    \STATE $b=b+1$
  \ENDFOR
\ENDFOR
\STATE $return \frac{w}{b}$
\end{algorithmic}
\end{algorithm}

\begin{algorithm}
\caption{Evaluate the interaction}
\label{alg:value}
\begin{algorithmic}[1] 
\REQUIRE $x_{1},x_{2}$
\COMMENT{two sensor value involved, for a more deeper tree, x1 can be an evaluation value from the previous level of depth}
\REQUIRE $l_{1},l_{2}$
\COMMENT{level of interaction}
\REQUIRE $a$
\COMMENT{weight of interaction}
\STATE $E=a((\frac{x_{1}}{l_{1}+1})+(\frac{x_{2}}{l_{2}+1}))$
\end{algorithmic}
\end{algorithm}

In positioning the sensors as determined by Algorithm \ref{alg:ps}, for the case of two sensors 
the algorithm only exchanges their sequences. After the evaluation of first sequence, the algorithm 
exchanges the sequence and reevaluate the new one. This procedure is specified in the second to fourth lines. 
For the case of more than two sensors, the algorithm recursively traverses the sequence
till it reaches the condition where only two sensors are left. In this case,
the algorithm runs in the same manner as if there are only two sensors involved. Each time
the algorithm is traversing deeper, the root is increased indicating the depth of tree under evaluation. 
If a leaf of tree is reached, the algorithm returns back to the parent leaf, exchanges to the next root, traverses  deeper till it reaches the leaf. After all paths to each leaf have been reached by one root, the sequence 
exchanges to another sensor as the new root and so forth.

Furthermore, the weight of interaction is calculated using Eq.
(\ref{eq:aij}) and given in Tab. \ref{tab:aij}. In Algorithm \ref{alg:ps} 
$root$ determines the $root$ of a new sub tree, while in Algorithm
\ref{alg:weight} $root$ is intended to determine the number of sensors which are currently
involved in the interaction as illustrated in Fig. \ref{fig:simpleEval}. $b$ in
Algorithm \ref{alg:weight} is intended to count the number of combination of two sensors 
among the whole sensors as defined in Eq. \ref{eq:aij2}, that is it would be as
many as elements in Tab. \ref{tab:aij}. For example, the interaction of three sensors 
contains three dual-sensor interactions, while the interaction of four sensors
contains six and so on. 

Finally, Algorithm \ref{alg:value} is intended to calculate the evaluation value, $E$. It
requires the weight of each interaction from Algorithm \ref{alg:weight} using Eq. 
(\ref{eq:aij}) and Tab. \ref{tab:aij}. The total number of evaluation values is 
equal to the number of sensors being involved. 

\section{Summary}

A new model based on the exhaustive search method for hybrid sensor network has
been proposed. The model treats all sensors in the same manner by introducing
normalization procedure for all sensors and parameters. It is shown that the
model is able to describe the whole evaluation processes using few parameters,
that is the coupling constant $a$ for each pair of sensors determined
uniformly and the universal threshold value $E_\mathrm{th}$.

In the present paper, the study is focused on the case of sensor network with
1-D relationship. A simple algorithm for such cases has also been given and
briefly discussed. Through the discussion, it is argued that the model could
realize a feasible early warning system for any safety critical facilities
involving various sensors using exhaustive method to prevent unnecessary
failures due to statistical approaches in, for example any AI-based methods. On
the other hand, the method requires much less computing power since it has only
a complexity of $O(n!)$. 

In principle the method can be extended to incorporate more complicated
relationship among the sensors by considering higher dimensional relationship.
Nevertheless, some studies on distributing the computation load to improve the
processing speed should also been done carefully. All of these issues are in
progress and will be published elsewhere.

\section*{Acknowledgment}

AAW thanks  the Indonesian Ministry of Research and Technology for financial
support, and appreciates the Group for Theoretical and Computational Physics at 
Research Center for Physics LIPI for warm hospitality during the work. The work
of LTH is supported by the Riset Kompetitif LIPI 2012 under Contract no. 
11.04/SK/KPPI/II/2012.

\bibliographystyle{IEEEtran}
\bibliography{icias2012}

\end{document}